\documentclass[conference]{IEEEtran}
\IEEEoverridecommandlockouts
\usepackage{cite}
\usepackage{amsmath,amssymb,amsfonts}
\usepackage{algorithmic}
\usepackage{graphicx}
\usepackage{textcomp}
\usepackage{xcolor}
\usepackage{url}
\def\BibTeX{{\rm B\kern-.05em{\sc i\kern-.025em b}\kern-.08em
    T\kern-.1667em\lower.7ex\hbox{E}\kern-.125emX}}
\begin{document}
\raggedbottom

\title{FaceNet versus VGG Features in Wasserstein Distortion for Face Image Compression}

\author{
\IEEEauthorblockN{Xinyu Wang}
\IEEEauthorblockA{
\textit{New York University}\\
New York, USA\\
xw3713@nyu.edu
}
\and
\IEEEauthorblockN{Jona Ball\'e}
\IEEEauthorblockA{
\textit{New York University}\\
New York, USA\\
jona.balle@nyu.edu
}
}

\maketitle

\begin{abstract}
Wasserstein distortion (WD) compares local statistics of deep features
and has been used to optimize learned image codecs for perceptual
quality, but existing WD objectives rely on features from networks
trained for generic image classification, such as VGG. We ask whether
features from a face recognition network better reflect the perceived
quality of compressed faces. In the Cool-Chic codec, we replace the VGG
features in a WD-based objective with intermediate feature maps of
FaceNet, and compare the two objectives on 20 LFW face images at a
matched average bitrate of about 0.15 bits per pixel. In a preliminary pairwise human study, FaceNet-based reconstructions were preferred in 72.5\% of
judgments and by a majority of participants on 16 of the 20 images. These results suggest that the
choice of feature network inside WD affects the perceived quality of
face images, and that face recognition features are a promising
direction for face-aware perceptual compression.
\end{abstract}

\begin{IEEEkeywords}
Learned Image Compression, Wasserstein Distortion, Visual Perception,
Face Recognition Features
\end{IEEEkeywords}

\section{Introduction}

Learned image compression aims to balance
bitrate and reconstruction quality through
rate-distortion optimization~\cite{balle2018hyperprior}.
The choice of distortion measure determines what the codec
preserves. Pixel-wise measures such as mean squared error (MSE) are simple to
optimize but agree poorly with human perception of image
quality~\cite{zhang2018lpips}; at low bitrates, MSE-optimized codecs
produce blurry reconstructions~\cite{blau2019rdp}. Perceptual
distortion measures are instead designed to agree with human
perception~\cite{zhang2018lpips,ding2022dists}, and have been used
to optimize codecs at low bitrates~\cite{mentzer2020hific}. Wasserstein distortion (WD)
is one such measure: rather than comparing images pixel by pixel, it
compares local statistics of visual feature
representations~\cite{qiu2024wasserstein,balle2025wasserstein}.

Facial regions pose a particular challenge for
perceptual image compression. Previous studies
have shown that the visibility of distortions
varies across facial regions and distortion
types~\cite{jiang2024facemap}.
Human face perception also involves sensitivity
to configural information, including the spatial
relationships among facial
features~\cite{maurer2002faces}.
These findings motivate closer attention to
facial reconstruction quality.
WD makes this question concrete. Because it only requires local
feature statistics to match, WD allows a codec to resynthesize
textures such as foliage or hair instead of transmitting them
exactly~\cite{balle2025wasserstein}. The choice of feature representation determines which reconstruction differences are measured by WD. This raises the question of whether features from networks trained for generic image classification adequately capture perceptually important differences in facial content.

VGG~\cite{simonyan2015vgg} was trained for generic image
classification, and its features are widely used for perceptual
image reconstruction~\cite{johnson2016perceptual}.
FaceNet~\cite{schroff2015facenet} was instead trained to tell
individuals apart, so its features are expected to be sensitive to
the shape and arrangement of facial parts. We therefore hypothesize
that, on face images, computing WD on FaceNet features rather than
VGG features makes the objective more sensitive to facial structure,
and yields reconstructions that human observers judge to be closer
to the original.
FaceNet identity embeddings have previously been
used as a training loss for a dedicated face
image codec~\cite{chen2019facial}.
Unlike this approach, we use the local statistics
of intermediate FaceNet feature maps within WD,
rather than distances between global identity embeddings.

As a first step, we compare two WD-based objectives
in the Cool-Chic codec~\cite{ladune2023coolchic}.
Both use the same image-level term, WD computation,
and codec configuration, but differ in the feature
network: a \emph{VGG-based} objective, which serves
as the baseline, and a \emph{FaceNet-based} objective,
which replaces VGG features with FaceNet features.
We evaluate both on 20 face-centered images from
LFW~\cite{huang2007lfw}, at closely matched average
bitrates, through a human pairwise preference study.

The main contributions of this work are as follows:
\begin{itemize}
  \item We investigate the use of intermediate face recognition
feature maps as an alternative representation to generic VGG
features within Wasserstein distortion for face image compression.
  \item In a human pairwise study on LFW face images at a matched
  average bitrate, the FaceNet-based objective is preferred in 72.5\% of
  judgments, consistently across participants and images.
  \item We characterize how the choice of feature representation
  within WD affects the perceptual quality of reconstructed faces.
\end{itemize}

\section{Related Work}
\label{sec:related}

\subsection{Learned and Perceptual Image Compression}
End-to-end optimized transform codecs~\cite{balle2017endtoend} with
hyperprior entropy models~\cite{balle2018hyperprior} now outperform
classical codecs in rate-distortion, but when trained for MSE they
produce blurry reconstructions at low bitrates. Blau and
Michaeli~\cite{blau2019rdp} formalized the resulting
rate-distortion-perception trade-off, and generative codecs use
adversarial and perceptual losses to synthesize plausible detail at
very low rates~\cite{agustsson2019extreme,mentzer2020hific}. A
complementary line of work overfits a lightweight decoder to each
image, as in Cool-Chic~\cite{ladune2023coolchic} and
C3~\cite{kim2024c3}, which reaches competitive rate-distortion with
very low decoding complexity. Wasserstein distortion
(WD)~\cite{qiu2024wasserstein} interpolates between fidelity and
realism by comparing local feature statistics within a spatially
varying pooling region. Ball\'e~\emph{et al.}~\cite{balle2025wasserstein}
optimized the overfitted codec C3 for WD on VGG
features~\cite{simonyan2015vgg} and showed, in a human rater study, that
WD clearly outperforms LPIPS as an optimization objective, reaching a
trade-off between visual quality and bitrate similar to generative codecs
such as HiFiC~\cite{mentzer2020hific} with less than 1\% 
of their decoding multiply-accumulate operations (MACs).
We build on this line of work, adapting the WD implementation in
Cool-Chic~5.0.0~\cite{ladune2026coolchic5} to investigate the effect
of the feature network inside WD.

\subsection{Feature-Based Perceptual Losses and Metrics}
Distances between deep features are now standard both as training
losses~\cite{johnson2016perceptual} and as full-reference quality
metrics. LPIPS~\cite{zhang2018lpips} compares normalized activations of
classification networks location by location, whereas
DISTS~\cite{ding2022dists} combines a structure term with a texture
term based on spatial means and correlations of VGG features, in the
spirit of Gram-matrix texture models~\cite{gatys2015texture}. WD is similar to DISTS in comparing feature statistics
rather than individual feature values, but performs
the comparison locally, with the pooling scale controlling
the trade-off between fidelity and texture resynthesis. All of
these methods rely on networks trained for generic object
recognition. We investigate whether a domain-specific face recognition network can provide a useful alternative feature representation within WD.

\subsection{Face Identity Features as Losses}
Face recognition networks such as FaceNet~\cite{schroff2015facenet}
and ArcFace~\cite{deng2019arcface} map faces to embeddings that are
discriminative of identity. Face hallucination and restoration methods have used
identity-preserving losses based on face recognition
features, as in SICNN~\cite{zhang2018sicnn} and
GFP-GAN~\cite{wang2021gfpgan}.
These losses compare a single global embedding of the face. Unlike ROI-based approaches, we do not use face masks
or explicitly modify bit allocation. Instead, we
investigate the effect of replacing VGG features
with face recognition features within WD.

\subsection{Face- and ROI-Aware Compression}
Region-of-interest (ROI) codecs allocate more bits to salient regions
given a mask, either through an explicit ROI
loss~\cite{akutsu2019roi,cai2020roi} or through a spatial quality map
that conditions a variable-rate model~\cite{song2021qmap}. For faces
specifically, Chen and He~\cite{chen2019facial} trained a learned codec
with a FaceNet embedding loss and gradient-driven spatial bit
allocation, and generative approaches transmit compact facial
representations and resynthesize the face at the decoder, from
keypoint-based talking-head video coding~\cite{wang2021oneshot} to
generative latent coding of face images at ultra-low
bitrates~\cite{jia2024glc}.

In contrast, we do not modify the codec architecture or explicitly
control spatial bit allocation: the only change is the
feature network inside the distortion measure.

\section{Proposed Method}
\label{sec:method}

Each image is compressed independently by optimizing a Cool-Chic
codec~\cite{ladune2023coolchic} to minimize the rate-distortion objective
\begin{equation}
\mathcal{L} = D(x,\hat{x}) + \lambda R,
\end{equation}
where $x$ and $\hat{x}$ denote the original and reconstructed images,
respectively, $R$ is the estimated bitrate, and $D$ is the perceptual
distortion. We investigate the effect of replacing generic image
classification features with face recognition features in a
Wasserstein-distortion-based objective, without modifying the codec's
decoder architecture.

\subsection{Wasserstein Distortion}
\label{sec:wd}

Wasserstein distortion (WD) compares visual representations
through pointwise differences and local feature
statistics~\cite{qiu2024wasserstein,balle2025wasserstein}.

Our experiments build upon the WD implementation in
Cool-Chic 5.0.0~\cite{ladune2026coolchic5}, which we extend with an
image-level term and the FaceNet features, using ImageNet-normalized
inputs and a smaller pooling scale. We denote by $\mathrm{WD}_\sigma(u,v)$ the Wasserstein
distortion between two tensors $u$ and $v$ at pooling scale $\sigma$,
following the definition in~\cite{qiu2024wasserstein,balle2025wasserstein}.
Due to the small size of the images in our study, we use a fixed $\log_2\sigma=0.5$ for the image and all feature tensors.
This small pooling scale places the objective close to the fidelity end
of WD, where local feature statistics are compared within $3\times3$
neighborhoods. Following this implementation, the pooling scale is
applied directly on each tensor, without rescaling it to the tensor's
resolution.

\subsection{Feature Representations}\label{sec:features}

\textbf{Image representation.}
Both objectives include an image-level WD term computed
from ImageNet-normalized RGB images. We denote this
shared contribution by $\mathrm{WD}_I$.

\textbf{VGG features.}
Following the Cool-Chic WD implementation, we use an ImageNet-pretrained
VGG-16~\cite{simonyan2015vgg} and extract activations from
\texttt{relu1\_2}, \texttt{relu2\_2}, \texttt{relu3\_3}, and
\texttt{relu4\_3} in a single forward pass on the ImageNet-normalized
image. We denote the four feature tensors by $V_1(x),\ldots,V_4(x)$.

\textbf{FaceNet features.}
We use an Inception-ResNet-v1~\cite{szegedy2017inception} face
recognition network pretrained on VGGFace2~\cite{cao2018vggface2}, as
implemented in \texttt{facenet-pytorch}, with inputs scaled to
$[-1,1]$. We extract four intermediate feature maps, from
\texttt{conv2d\_2b} (64 channels), \texttt{conv2d\_4b} (256),
\texttt{repeat\_1} (256), and \texttt{repeat\_2} (896), and denote them
by $F_1(x),\ldots,F_4(x)$.

\subsection{Distortion Objectives}\label{sec:objectives}

Let $N(x)$ denote the ImageNet-normalized image.
The shared image-level distortion is defined as
\begin{equation}
\mathrm{WD}_I(x,\hat{x})
=
\mathrm{WD}_\sigma\bigl(N(x),N(\hat{x})\bigr).
\end{equation}

We define the VGG- and FaceNet-based feature
distortions, respectively, as
\begin{equation}
\mathrm{WD}_V(x,\hat{x})
=
\sum_{k=1}^{4}
\mathrm{WD}_\sigma
\bigl(V_k(x),V_k(\hat{x})\bigr),
\end{equation}
\begin{equation}
\mathrm{WD}_F(x,\hat{x})
=
\sum_{k=1}^{4}
\mathrm{WD}_\sigma
\bigl(F_k(x),F_k(\hat{x})\bigr).
\end{equation}

The two distortion objectives are then
\begin{equation}
\begin{aligned}
D_V &= \mathrm{WD}_I + \mathrm{WD}_V,\\
D_F &= \mathrm{WD}_I + \mathrm{WD}_F.
\end{aligned}
\end{equation}

The VGG-based objective $D_V$ serves as the baseline.

\section{Experimental Setup}
\label{sec:setup}

\subsection{Datasets}

We evaluate the two objectives on 20 face images randomly
sampled from Labeled Faces in the Wild (LFW)~\cite{huang2007lfw}.
The images are selected from
the original, unaligned LFW dataset using a fixed
random seed of 0, with each image representing a
different identity. Grayscale images are excluded.

All images have a resolution of $250 \times 250$
pixels. They are converted to RGB PNG format without
additional cropping or resizing.

\subsection{Compression Configurations}

We use Cool-Chic~5.0.0~\cite{ladune2023coolchic,ladune2026coolchic5},
optimizing the codec independently for each image over 10{,}000
iterations with the \texttt{mop} (medium operating point) decoder
configuration, which requires about 1000 multiplications per decoded
pixel. Both objectives use the same codec configuration, and the
distortion consists only of the WD terms defined in
Section~\ref{sec:objectives}; we do not use the MSE term that Cool-Chic
combines with WD in its default perceptual setting.

\subsection{Bitrate Control}

We target an average bitrate of approximately
$0.15$ bits per pixel (bpp) across the 20 images.

Because $\mathrm{WD}_V$ and $\mathrm{WD}_F$ differ in magnitude,
the two objectives require different values of $\lambda$ to reach
the same bitrate. For the FaceNet-based objective, we use
$\lambda=0.01$ for all images. For the VGG-based
objective, we select image-specific values from
$\lambda\in\{0.24,0.25,0.30,0.36\}$ to obtain
a comparable aggregate bitrate.

The average coded bitrates, computed from the bitstream sizes and
matched over the dataset rather than per image, are 0.14726~bpp for
FaceNet and 0.14727~bpp for VGG, so the FaceNet-based reconstructions
do not benefit from a higher bitrate.

\subsection{Human Perceptual Evaluation}

We conduct a human perceptual study using the Mabyduck
platform~\cite{mabyduck}. Participants compare
VGG-based and FaceNet-based reconstructions of the
same reference image using a flip-between interface,
which allows them to toggle between the two
reconstructions.

The reference image is displayed, and participants
answer the question: ``Which image looks more similar
to the reference?'' Ties, zooming, and panning are
disabled. Participants are not told which method produced which
reconstruction, the assignment of the two reconstructions to the toggle
states is randomized by the platform, and the order of image pairs is
selected by Mabyduck's information-gain maximization strategy.

Eight volunteers participated, all of whom were either members of our research group or acquaintances of the first author. Two of the participants were authors of this paper.
Each participant judged all 20 image pairs, giving
160 pairwise judgments in total. In each trial, the reference
image, the VGG-based reconstruction, and the FaceNet-based
reconstruction of the same image are shown.
Because the judgments share participants and images, they are not
independent. We therefore quantify uncertainty mainly with a bootstrap
that resamples both participants and images, and complement it with
sign tests over images and over participants. We also report the
Bayesian Elo scores estimated by the platform.

\section{Results and Discussion}
\label{sec:results}

\subsection{Human Perceptual Preferences}

\begin{figure}[t]
\centering
\includegraphics[width=\linewidth]{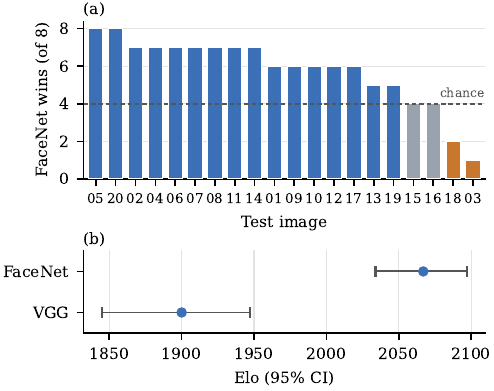}
\caption{Human pairwise study on the 20 LFW test images (about
0.15~bpp). (a)~Number of the eight participants who chose the
FaceNet-based reconstruction over the VGG-based one, per image; the
dashed line marks chance (4 of 8). Images are sorted by FaceNet wins.
(b)~Bayesian Elo scores (posterior means of a Bradley--Terry model)
with 95\% confidence intervals, as estimated by the Mabyduck platform.}
\label{fig:votes}
\end{figure}

Figure~\ref{fig:votes} presents the results of the human perceptual
evaluation: panel~(a) shows the per-image votes and panel~(b) the Elo scores.
The FaceNet-based reconstruction was chosen in 116 of 160 judgments
(72.5\%; 95\% bootstrap interval over participants and images:
[56.9\%, 87.5\%]). The preference is consistent across images: FaceNet
was preferred by most participants on 16 of the 20 images and VGG on 2,
with 2 ties (sign test, $p=0.001$). Seven of eight participants
preferred FaceNet (sign test, $p=0.07$). Excluding the two participants
who are authors, FaceNet was chosen in 84 of 120 judgments (70.0\%).

The Bayesian Elo scores estimated by the platform agree with these
results: 2067 (95\% CI [2034, 2097]) for FaceNet and 1900
(95\% CI [1845, 1947]) for VGG, corresponding to a preference probability of
0.72. These intervals assume independent judgments and are therefore
narrower than the bootstrap interval.
These results indicate that, on this set of face-centered images and
at this operating point, human observers judged FaceNet-based
reconstructions to be closer to the reference more often than
VGG-based ones. Given the size of the study, we regard this as
preliminary evidence that has yet to be confirmed on larger and more
diverse image sets.

\subsection{Qualitative Comparison}

\begin{figure}[t]
\centering
\includegraphics[width=\linewidth]{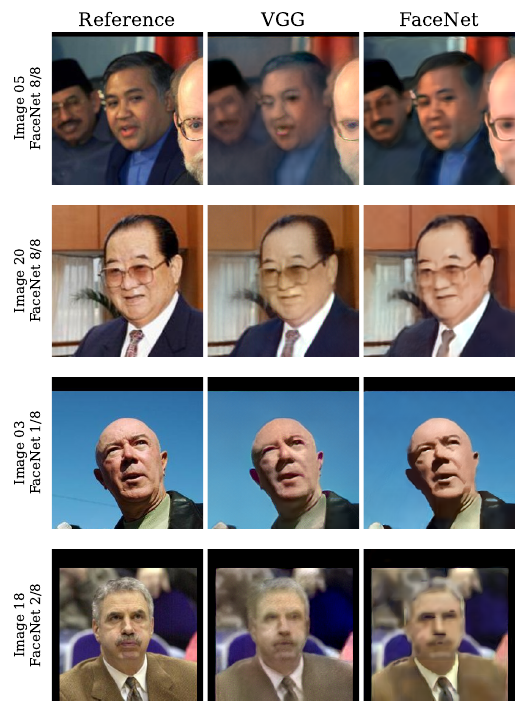}
\caption{Reconstructions of four test images. Images 05 and 20: FaceNet
preferred by all eight participants. Images 03 and 18: the two images on
which VGG was preferred by most participants. Best viewed zoomed in.}
\label{fig:qual}
\end{figure}

Figure~\ref{fig:qual} shows two images on which all participants
preferred the FaceNet-based reconstruction (05 and 20) and the two
images on which most participants preferred the VGG-based one (03 and
18). In the VGG-based reconstructions of images 05 and 20, facial
features are blurred and partly doubled: in image 05, the eyes and mouth
of the central face lose their shape, and in image 20, the frames of the
glasses and the eyes behind them become faint. Both also appear less
saturated than the reference. The FaceNet-based reconstructions keep
sharper outlines of the eyes, eyebrows, mouth, and glasses, including on
the smaller faces at the edges of image 05, although fine skin texture
such as wrinkles is smoothed away.

The two failure cases show the opposite tendency. In image 03, the
facial contours are similar in both reconstructions, but the
FaceNet-based one looks smoothed and loses most of the skin texture,
whereas the VGG-based one is closer to the reference in color. In
image 18, the FaceNet-based reconstruction replaces the eyes with dark
horizontal bars and renders the mustache and face outline blocky,
whereas the VGG-based one is blurrier but keeps plausible facial
features. Together, these examples suggest that the FaceNet-based
objective better preserves facial contours, but can lose skin texture
and introduce blocky artifacts. 

\subsection{Discussion}

The feature representation used within WD determines which
reconstruction differences are penalized during optimization, and may
therefore affect which visual properties the codec preserves. Our
results suggest that this choice is visible to human observers. With
VGG features, facial contours can drift: eyes, mouth, and glasses
become blurred or doubled. With FaceNet features, facial contours
remain sharper, but skin texture is less well preserved, and some
images show blocky artifacts. The overall preference for FaceNet
suggests that, on most of our test images, participants found the loss
of facial contours more noticeable than the loss of skin texture. The
failure cases, however, show that neither representation captures
everything that matters for faces. The two representations may thus
capture different perceptually relevant properties, which motivates
investigating whether objectives that combine face-specific and
generic features can benefit from both.

\subsection{Limitations}

\emph{Scope.} Our evaluation is limited to 20 face-centered LFW images,
a single operating point at an average bitrate of about 0.15~bpp,
one face recognition network, and a modified WD implementation based
on Cool-Chic~5.0.0 with a fixed pooling parameter of
$\log_2\sigma=0.5$. The generalizability of our findings beyond
these experimental conditions remains to be established.

\emph{Experimental control.} The terms are summed with unit weights and
are not normalized, so their relative contributions differ between the
two objectives. The observed preference may therefore partly reflect
this weighting rather than the feature content alone.

\emph{Study size.} The perceptual study involves eight participants,
two of whom are authors, and 20 images. Larger studies with
participants unaffiliated with the research team and a broader
image set are needed to assess the generalizability of our findings.

\section{Conclusion}
\label{sec:conclusion}

We investigated the use of face recognition features
as an alternative to generic VGG features within a
Wasserstein-distortion-based objective for learned
image compression.

Using Cool-Chic, we compared VGG- and FaceNet-based
objectives on 20 LFW face images at closely matched
average bitrates. Human perceptual evaluation favored
the FaceNet-based objective: participants chose the
FaceNet-based reconstructions in 72.5\% of judgments.

These findings provide preliminary evidence that the
choice of feature network inside WD affects the perceived
quality of compressed face images. The observed differences
in reconstruction artifacts suggest that the two
representations may capture different perceptually relevant
properties, motivating further investigation of their
potential complementarity.

Building on this, our ongoing work examines whether
face-specific representations remain useful in more
complex images where faces occupy only part of the scene,
where they fail, and how they can be combined with generic
visual features within a single objective, using an updated
WD implementation, larger pooling scales, and multiple
operating points.

\bibliographystyle{IEEEtran}
\bibliography{references}

\end{document}